\pdfoutput=1

\documentclass[11pt]{article}

\usepackage{EMNLP2023}

\usepackage{times}                            
\usepackage{latexsym}                         
\usepackage[T1]{fontenc}
\usepackage[utf8]{inputenc} 
\usepackage{microtype}                        
\usepackage{inconsolata}                      
\usepackage{booktabs}                         
\usepackage[raggedrightboxes]{ragged2e}       
\usepackage{subfiles}                         
\usepackage{longtable}                        
\usepackage{amsmath}                          
\usepackage{graphicx}                         

\title{calamanCy: A Tagalog Natural Language Processing Toolkit}

\author{Lester James V. Miranda \\
  ExplosionAI GmbH \\
  \texttt{lj@explosion.ai}}

\begin{document}
\maketitle
\begin{abstract}
  We introduce calamanCy, an open-source toolkit for constructing natural language processing (NLP) pipelines for Tagalog.
  It is built on top of spaCy, enabling easy experimentation and integration with other frameworks.  
  calamanCy addresses the development gap by providing a consistent API for building NLP applications and offering general-purpose multitask models with out-of-the-box support for dependency parsing, parts-of-speech (POS) tagging, and named entity recognition (NER).
  calamanCy aims to accelerate the progress of Tagalog NLP by consolidating disjointed resources in a unified framework.
  The calamanCy toolkit is available on GitHub: \url{https://github.com/ljvmiranda921/calamanCy}.
\end{abstract}

\section{Introduction}

Tagalog is a low-resource language from the Austronesian family, with over 28 million speakers in the Philippines \citep{Lewis2009EthnologueL}.
Despite its speaker population, few resources exist for the language \citep{Cruz2021ImprovingLL}. 
For example, Universal Dependencies (UD) treebanks for Tagalog are tiny ($\ll$ 20k words) \citep{Samson2018TRG,Aquino2020ParsingIT}, 
while domain-specific corpora are sparse \citep{Cabasag2016HatespeechIP,Livelo2018IntelligentDI}. 
In addition, Tagalog language models (LMs) \citep{Cruz2021ImprovingLL,Jiang2021PretrainedLM} are few, while most multilingual LMs \citep{Conneau2019UnsupervisedCR,Devlin2019BERTPO} underrepresent the language \citep{Lauscher2020FromZT}.
Thus, consolidating these disjointed resources in a coherent framework is still an open problem.
The lack of such framework hampers model development, experimental workflows, and the overall advancement of Tagalog NLP.

To address this problem, we introduce calamanCy,\footnote[1]{
  ``calamanCy'' derives its name from \textit{kalamansi}, a citrus fruit native to the Philippines.}
an open-source toolkit for Tagalog NLP. 
It is built on top of spaCy \citep{Honnibal2020Spacy} and offers end-to-end pipelines for NLP tasks such as dependency parsing, parts-of-speech (POS) tagging, and named entity recognition (NER). 
calamanCy also provides general-purpose pipelines in three different sizes to fit any performance or accuracy requirements.
This work has two main contributions: (1) an open-source toolkit with out-of-the box support for common NLP tasks, and (2) comprehensive evaluations on several Tagalog benchmarks.

\begin{table*}[t]
\begin{tabular}{@{}p{2cm}p{9cm}p{4cm}@{}}
\toprule
Entity             & Description                                                                                                                    & Examples                                       \\ \midrule
Person (PER)       & Person entities limited to humans. It may be a single individual or group.                                                     & Juan de la Cruz, Jose Rizal, Quijano de Manila \\
Organization (ORG) & Organization entities limited to corporations, agencies, and other groups of people defined by an organizational structure.    & Meralco, DPWH, United Nations                  \\
Location (LOC)     & Location entities are geographical regions, areas, and landmasses. Geo-political entities are also included within this group. & Pilipinas, Manila, CALABARZON, Ilog Pasig        \\ \bottomrule
\end{tabular}
\caption{
Entity types used for annotating TLUnified-NER (derived from the TLUnified pretraining corpus of \citealp{Cruz2021ImprovingLL}).
}
\label{table:entity_types}
\end{table*}

\section{Related Work}

\paragraph*{Open-source toolkits for NLP}
There has been a growing body of work in the development of NLP toolkits in recent years.
For example, DaCy \citep{Enevoldsen2021DaCyAU} and HuSpaCy \citep{Orosz2022HuSpaCyAI} serve the language-specific needs of Danish and Hungarian respectively.
In addition, scispaCy \citep{Neumann2019ScispaCyFA} and medspaCy \citep{Eyre2021LaunchingIC} were built to focus on scientific text.
These tools employ spaCy \citep{Honnibal2020Spacy}, an industrial-strength open-source software for natural language processing.
Using spaCy as a foundation is optimal, given its popularity and integration with other frameworks such as HuggingFace transformers \citep{Wolf2019HuggingFacesTS}.
However, no tool has existed for Tagalog until now.
We aim to fill this development gap and serve the needs of the Tagalog language community through calamanCy.

\paragraph*{Evaluations on Tagalog NLP Tasks} 
Structured evaluations for core NLP tasks, such as dependency parsing, POS tagging, and NER, are meager.
However, we have access to a reasonable amount of data to conduct comprehensive benchmarks.
For example, TLUnified \citep{Cruz2021ImprovingLL} is a pretraining corpus that combines news reports \citep{Cruz2020ExploitingNA}, a preprocessed version of CommonCrawl \citep{OrtizSuarez2019AsynchronousPF}, and several other datasets.
However, it was evaluated on domain-specific corpora that may not easily transfer to more general tasks.
In addition, Tagalog has two Universal Dependencies (UD) treebanks, Tagalog Reference Grammar (TRG) \citep{Samson2018TRG} and Ugnayan \citep{Aquino2020ParsingIT}, both with POS tags and relational structures for parsing grammar.
This paper will fill the evaluation gap by providing structured benchmarks on these core tasks.

\section{Implementation}

The best way to use calamanCy is through its trained pipelines.
After installing the library, users can access the models in a few lines of code:

\begin{verbatim}
  import calamancy as cl
  nlp = cl.load("tl_calamancy_md-0.1.0")
  doc = nlp("Ako si Juan de la Cruz.")
\end{verbatim}

Here, the variable \texttt{nlp} is a spaCy processing pipeline\footnote[2]{\url{https://spacy.io/usage/processing-pipelines}} that contains trained components for POS tagging, dependency parsing, and NER.
Applying this pipeline to a text will produce a \texttt{Doc} object with various linguistic features. 
calamanCy offers three pipelines of varying capacity: two static word vector-based models (md, lg), and one transformer-based model (trf).
We will discuss how we developed these pipelines in the following section.

\subsection{Pipeline development}

\paragraph*{Data annotation for NER}
There is no gold-standard corpus for NER, so we built one. 
To construct the NER corpus, we curated a portion of TLUnified \citep{Cruz2021ImprovingLL} to contain Tagalog news articles.
Including the author, we recruited two more annotators with at least a bachelor's degree and whose native language is Tagalog.
The three annotators labeled for four months, given three entity types as seen in Table \ref{table:entity_types}.
We chose the entity types to resemble ConLL \citep{Sang2002IntroductionTT,Sang2003IntroductionTT}, a standard NER benchmark.
We excluded the \texttt{MISC} label to reduce uncertainty and confusion when labeling.
Then, we measured inter-annotator agreement (IAA) by taking the pairwise Cohen's $\kappa$ on all tokens and then averaged them for all three pairs.
This process resulted in a Cohen's $\kappa$ score of 0.81. 
To avoid confusion with the original TLUnified pretraining corpora, we will refer to this annotated NER dataset as TLUnified-NER.
The final dataset statistics can be found in Table \ref{table:dset_stats}.
For the dependency parser and POS tagger, we merged the TRG \citep{Samson2018TRG} and Ugnayan \citep{Aquino2020ParsingIT} treebanks to leverage their small yet relevant examples.

\begin{table}[t]
\begin{tabular}{@{}lllll@{}}
\toprule
Dataset     & Examples & PER  & ORG  & LOC  \\ \midrule
Training    & 6252     & 6418 & 3121 & 3296 \\
Development & 782      & 793  & 392  & 409  \\
Test        & 782      & 818  & 423  & 438  \\ \bottomrule
\end{tabular}
\caption{Dataset statistics for TLUnified-NER.}
\label{table:dset_stats}
\end{table}

\begin{table*}[t]
\begin{tabular}{@{}p{3cm}p{4cm}p{4cm}p{3.75cm}@{}}
\toprule
Pipeline  &  Pretraining objective & Word embeddings & Dimensions \\ \midrule
Medium-sized pipeline (tl\_calamancy\_md) & Predict some number of leading and trailing UTF-8 bytes for the words. & Uses floret vectors trained on the TLUnified corpora. & 50k unique vectors (200 dimensions), Size: 77 MB\\
Large-sized pipeline (tl\_calamancy\_lg)  & Same pretraining objective as the medium-sized pipeline.    & Uses fastText vectors trained on CommonCrawl corpora.                  & 714k unique vectors (300 dimensions), Size: 455 MB \\
Transformer-based pipeline (tl\_calamancy\_trf) & No separate pretraining because there's no token-to-vector component. & Context-sensitive vectors from a transformer network.      & Uses roberta-tagalog-base. Size: 813 MB  \\ \bottomrule
\end{tabular}
\caption{
    Language pipelines available in calamanCy (v0.1.0).
    The pretraining method for the word-vector models is a variant of the \textit{cloze task}.
    All pipelines have a \texttt{tagger}, \texttt{parser}, \texttt{morphologizer}, and \texttt{ner} spaCy component.
}
\label{table:calamancy_pipelines}
\end{table*}

\begin{table*}[t]
\begin{tabular}{@{}p{4cm}p{4cm}p{7.25cm}@{}}
\toprule
Dataset &  Task / Labels & Description  \\ \midrule
Hatespeech \citep{Cabasag2016HatespeechIP} & Binary text classification \textit{(hate speech, not hate speech)} & Contains 10k tweets collected during the 2016 Philippine Presidential Elections labeled as hate speech or non-hate speech. \\
Dengue \citep{Livelo2018IntelligentDI}  & Multilabel text classification \textit{(absent, dengue, health, sick, mosquito)}   & Contains 4k dengue-related tweets collected for a health infoveillance application that classifies text into dengue subtopics.\\
TLUnified-NER \citep{Cruz2021ImprovingLL} & Named entity recognition \textit{(Person, Organization, Location)} & A held-out test split from the annotated TLUnified corpora containing news reports and other articles. See Table \ref{table:dset_stats}. \\ 
Merged UD \citep{Samson2018TRG,Aquino2020ParsingIT} & Dependency parsing and POS tagging & Merged version of the Ugnayan and TRG treebanks from the Universal Dependencies framework. \\ \bottomrule
\end{tabular}
\caption{
    Datasets for benchmarking calamanCy. 
}
\label{table:benchmark_datasets}
\end{table*}

\paragraph*{Model training}

We considered three design dimensions when training the calamanCy pipelines: (1) the presence of pretraining, (2) the word representation, and its (3) size or dimension.
Model \textit{pretraining} involves learning vectors from raw text to inform model initialization.
Here, the pretraining objective asks the model to predict some number of leading and trailing UTF-8 bytes for the words\textemdash a variant of the cloze task \citep{Devlin2019BERTPO}.
A model's \textit{word representation} may involve training static word embeddings using floret,\footnote[3]{\url{https://github.com/explosion/floret}} an efficient version of fastText \citep{Bojanowski2016EnrichingWV}, or using context-sensitive vectors from a transformer \citep{Vaswani2017AttentionIA}.
Finally, a model's \textit{dimension} is our way to tune the tradeoff between performance and accuracy.

The general process involves pretraining a filtered version of TLUnified, constructing static word embeddings if necessary, and training the downstream components.
We used TLUnified-NER to train the NER component, and then trained the dependency parser and POS tagger using the combined treebanks.
Ultimately, we devised three language pipelines as seen in Table \ref{table:calamancy_pipelines}.

\section{Evaluation}

\begin{table*}[t]
\begin{tabular}{@{}p{4cm}p{2cm}p{2cm}p{2cm}p{2cm}p{2cm}@{}}
\toprule
                           & \multicolumn{2}{c}{\textbf{Text categorization}} & \textbf{NER} & \multicolumn{2}{c}{\textbf{Dep. pars. \& POS tag.}}                         \\ 
\textbf{Model}             & Hatespeech (binary) & Dengue (multilabel) & TLUnified-NER & Merged UD, UAS~/~LAS & Merged UD, POS Acc. \\ \midrule 
\textit{Monolingual (Ours)}              \\
tl\_calamancy\_md          & 74.40$\pm$0.05 & 65.32$\pm$0.04 & 87.67$\pm$0.03 & 76.47~/~54.40 & 96.70\\
tl\_calamancy\_lg          & 75.62$\pm$0.02 & 68.42$\pm$0.01 & 88.90$\pm$0.01 & 82.13~/~70.32 & 97.20\\
tl\_calamancy\_trf         & \textbf{78.25$\pm$0.06} & \textbf{72.45$\pm$0.02} & \textbf{90.34$\pm$0.02} & \textbf{92.48~/~80.90} & \textbf{97.80} \\ \midrule
\textit{Cross-lingual transfer} \\
uk\_core\_news\_trf        & 75.24$\pm$0.03 & 65.57$\pm$0.01 & 51.11$\pm$0.02 & 54.77~/~37.68 & 82.86 \\
ro\_core\_news\_lg         & 69.01$\pm$0.01 & 59.10$\pm$0.01 & 02.01$\pm$0.00 & 84.65~/~65.30 & 82.80 \\
ca\_core\_news\_trf        & 70.01$\pm$0.02 & 59.42$\pm$0.03 & 14.58$\pm$0.02 & 91.17~/~79.30 & 83.09 \\ \midrule
\textit{Multilingual finetuning} \\
xlm-roberta-base          & 77.57$\pm$0.01 & 67.20$\pm$0.01 & 88.03$\pm$0.03 & 88.34~/~76.07 & 94.29 \\
bert-base-multilingual    & 76.40$\pm$0.02 & 71.07$\pm$0.04 & 87.40$\pm$0.02 & 90.79~/~78.52 & 95.30 \\
\bottomrule
\end{tabular}
\caption{
    Benchmark evaluation scores for monolingual, cross-lingual, and multilingual pipelines across a variety of tasks and datasets.
    We evaluated the text categorization and NER tasks across five trials, and then conducted 10-fold cross-validation for dependency parsing.
    F1-scores are reported on the text categorization and NER tasks.
}
\label{table:results}
\end{table*}

\paragraph*{Architectures}

We used spaCy's built-in architectures for each component in the calamanCy pipeline.
The token-to-vector layer uses the multi-hash embedding trick \citep{Miranda2022MultiHE} to reduce the representation size.
For the parser and named entity recognizer, we used a transition-based parser that maps text representations into a series of state transitions.
As for the text categorizer, we utilized an ensemble of a bag-of-words model and a feed-forward network.

\paragraph*{Experimental set-up} 
We assessed the calamanCy pipelines on various Tagalog benchmarks as detailed in Table \ref{table:benchmark_datasets}.
We also tested on text categorization, an unseen task, for robustness.
For NER evaluation, we used a held-out test split from TLUnified-NER.
We measured their performance across five trials and then reported the average and standard deviation.
For treebank-related benchmarks (POS tagging and dependency parsing), we followed UD's data split guidelines \citep{Nivre2020UniversalDV} and performed 10-fold cross-validation to compensate for the size of the corpora ($\ll$ 20k tokens).

We also tested a cross-lingual transfer learning approach, i.e., finetuning a model from a source language closely related to Tagalog.
According to \citet{Aquino2020ParsingIT}, the closest languages to Tagalog are Indonesian (id), Ukrainian (uk), Vietnamese (vi), Romanian (ro), and Catalan (ca).
They obtained these results via a distance metric \citep{Agic2017CrossLingualPS} based on the World Atlas for Language Structures \citep{Haspelmath2005WALS}.
However, only uk, ro, and ca have equivalent spaCy pipelines, so we only compared against those three.
Finally, we also compared against multilingual language models
by finetuning on XLM RoBERTa \citep{Conneau2019UnsupervisedCR} and an uncased version of multilingual BERT \citep{Devlin2019BERTPO}.
These LMs contain Tagalog in their training pool and are common alternatives for building Tagalog NLP applications.

\section{Discussion}

Table \ref{table:results} shows the F1-scores for the text categorization and NER tasks, the unlabeled (UAS) and labeled attachment scores (LAS) for the dependency parsing task, and the tag accuracy for POS tagging.

The calamanCy pipelines are competitive across all core NLP tasks while maintaining a smaller compute footprint.
As shown in the text categorization and NER results, users with low compute budgets can attain similar performance to multilingual LMs by using medium- or large-sized calamanCy models.
The transformer-based calamanCy pipeline is the best option for users who prioritize accuracy.
However, we were surprised that most alternative approaches perform better in dependency parsing.
We attribute this performance to the added strength of multilingual and cross-lingual information, which we don't have when training solely on a smaller treebank.
We plan to improve dependency parsing performance by building a larger treebank within the Universal Dependencies framework.
For practical applications, we recommend users to start with a medium- or large-sized calamanCy model before trying out GPU-intensive pipelines. 
Only then can they switch to a transformer-based pipeline to get accuracy gains.

\section{Conclusion}

In this paper, we introduced calamanCy, a natural language processing toolkit for Tagalog.
Our work has two main contributions: (1) an open-source toolkit containing general-purpose multitask pipelines with out-of-the-box support for common NLP tasks, and
(2) comprehensive benchmarks that compare against alternative approaches, such as cross-lingual or multilingual finetuning. 
We hope that calamanCy is a step forward to improving the state of Tagalog NLP. 
As a low-resource language, consolidating resources into a unified framework is crucial to advance research and improve collaboration.
In the future, we plan to create a more fine-grained NER benchmark corpus and extend calamanCy to natural language understanding (NLU) tasks.
Finally, the project is hosted on GitHub (\url{https://github.com/ljvmiranda921/calamanCy}) and we are happy to receive community feedback and contributions.

\section*{Limitations}

The TLUnified-NER corpus utilized for training the NER component of calamanCy comprises of new articles from early 2000s to the present.
In addition the Universal Dependencies (UD) corpora for the POS tagger and dependency parser components are relatively modest in size, containing fewer than 10k tokens.
Hence, the performance for these tasks during test-time could potentially be constrained by these factors.

Finally, reproducing the transformer pipelines may require a T4 or V100 GPU. 
The biggest bottleneck for reproduction is pretraining on the whole TLUnified corpus.
In a 64vCPU machine with 256GB of RAM, the pretraining process can take three full days for 20 epochs.

\bibliography{custom}
\bibliographystyle{acl_natbib}

\appendix

\section{Appendix}

\subsection{Reproducibility}

All the experiments and models in this paper are available publicly. 
Readers can head over to \url{https://github.com/ljvmiranda921/calamanCy} for all related software.
Note that the XLM-RoBERTa and multilingual BERT experiments may at least require a T4 or V100 GPU.

To reproduce the calamanCy models, head over to \texttt{models/v0.1.0}.
To reproduce the benchmarking experiments, head over to the \texttt{report/benchmark} directory.
Readers who are interested in the training set-up (e.g., hyperparameters, architectures used, etc.) can check the configuration (\texttt{.cfg}) files in the respective project's \texttt{configs/} directory.

\subsection{Building the TLUnified-NER corpus}

\begin{figure}[t]
\centering
\includegraphics[width=0.5\textwidth]{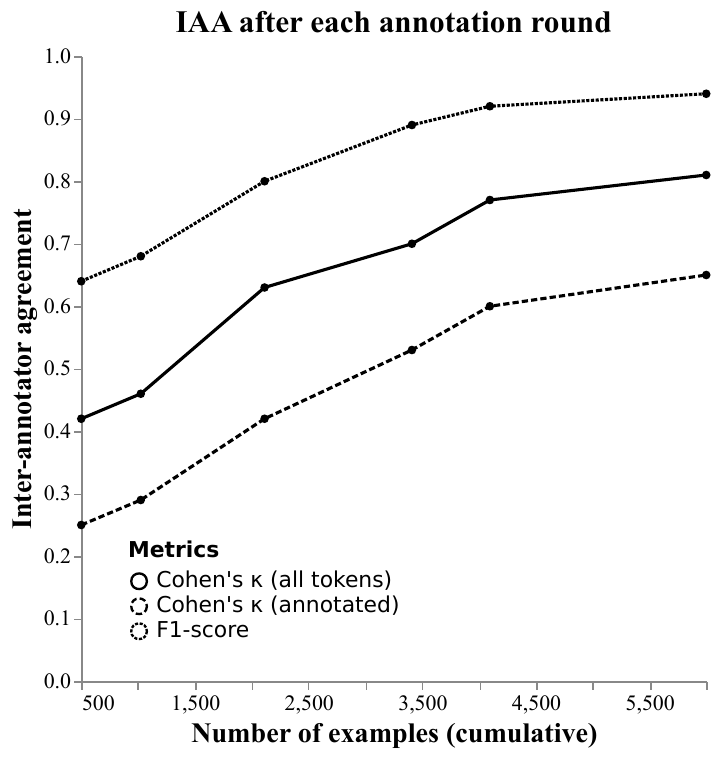}
\caption{
  Inter-annotator agreement measurement after each annotation round.
  Each mark represents the end of a round. 
  For each round, the annotators discuss disagreements, update the annotation guidelines, and evaluate the current set of annotations.
}
\label{fig:iaa}
\end{figure}

The TLUnified-NER dataset is a named entity recognition corpus containing the \textit{Person (PER)}, \textit{Organization (ORG)}, and \textit{Location  (LOC)} entities.
It includes news articles and other texts in Tagalog from 2009 to 2020.
It was based on the TLUnified pretraining corpora by \cite{Cruz2021ImprovingLL}.
The author, together with two more annotators, annotated TLUnified in the course of four months.
We employed an iterative approach as recommended by \citet{Reiter2017HT}, which included resolving disagreements and updating the annotation guidelines.

To get the inter-annotator agreement (IAA) score, we took \citet{Brandsen2020CreatingAD}'s work on the Archaeology dataset as inspiration.
We computed Cohen's $\kappa$ for all tokens, and only annotated tokens.  
In addition, we also measured the (3) pairwise F1 score without the `O' label \citep{Deleger2012BG}.
Table \ref{table:iaa} shows the IAA measurements while Figure \ref{fig:iaa} shows their growth after each annotation round.

\begin{table}[t]
\begin{tabular}{@{}p{6cm}p{1.25cm}@{}}
\toprule
Metric &  IAA  \\ \midrule
Cohen's $\kappa$ on all tokens & 0.81 \\ 
Cohen's $\kappa$ on annotated tokens only & 0.65 \\
F1 score & 0.91 \\ \bottomrule
\end{tabular}
\caption{
    Inter-annotator agreement (IAA) measurements.
    We obtained these values by computing for the pairwise comparisons between all annotator-pairs and averaging the results.
}
\label{table:iaa}
\end{table}

\end{document}